\documentclass[lettersize, conference]{class/ieeeconf}  

\IEEEoverridecommandlockouts                              
\usepackage{lipsum}
\usepackage{graphicx}      
\usepackage{float}
\usepackage[maxfloats=256]{morefloats}
\usepackage{tikz}
\usepackage{adjustbox}
\usepackage{pgfplots}
\usepgfplotslibrary{groupplots}
\usetikzlibrary{pgfplots.groupplots}
\usetikzlibrary{plotmarks}
\usetikzlibrary{calc}
\usepgfplotslibrary{external}
\pgfplotsset{compat=newest}

\usetikzlibrary{matrix}
\pgfplotsset{grid style={dashed,gray}}
 \usepackage{overpic}
 \graphicspath{{./}{graphics/}}
\newlength{\figureheight}
\newlength{\figurewidth}
\usepackage{color}
\usepackage{amsmath} 
\usepackage{amssymb}  
\usepackage{bm}  
\usepackage{makecell}
\usepackage{array}
\usepackage{arydshln}
\usepackage{nicematrix}
\usepackage{siunitx}
\usepackage{subcaption}
\usepackage{lpic}
\usepackage{flushend}
\usepackage{makecell}

\usepackage[T1]{fontenc}
\usepackage{hyperref}
\hypersetup{
    colorlinks=true,
    linkcolor=blue,
    filecolor=magenta,      
    urlcolor=cyan,
}

\usepackage{xspace}
\providecommand{\kuka}{\textsc{KUKA} LBR iiwa R820\xspace}
\providecommand{\schunk}{\textsc{SCHUNK} Dextrous Hand 2.0\xspace}
\providecommand{\camera}{\textsc{Basler} AVA 1000-100GC\xspace}
\providecommand{\esym}{\mathrm{e}}
\providecommand{\csym}{\mathrm{c}}
\providecommand{\hsym}{\mathrm{h}}
\providecommand{\osym}{\mathrm{o}}
\providecommand{\gsym}{\mathrm{g}}
\providecommand{\psym}{\mathrm{p}}

\usepackage{mathtools,amssymb}
\newcommand{\transpose}[1]{#1^\mathrm{T}}
\renewcommand{\vec}[1]{\mathbf{#1}}
\providecommand{\FullStop}{\text{~\@.\xspace}}
\providecommand{\Comma}{\text{~,\xspace}}

\newcommand{\braces}[1]{\left(#1\right)}
\newcommand{\rbraces}[1]{\left[#1\right]}

\newcommand{\PartDiff}[2]{\frac{\partial #1}{\partial #2}}

\usepackage{balance}

\title{\LARGE \bf
Online Trajectory Replanner for Dynamically Grasping \\Irregular Objects}

    \author{Minh Nhat Vu$^{1}$, Florian Grander$^{2}$, Anh Nguyen$^{3}$ 
    \thanks{$^{1}$ Minh Nhat Vu, is with the Automation \& Control Institute (ACIN), TU Wien, 1040 Vienna, Austria
            {\tt\small vu@acin.tuwien.ac.at}}%
    \thanks{$^{2}$ Florian Grander is with Automation and Control Center, Egger GmbH, 
           {\tt\small florian.grander@egger.com}}%
    \thanks{$^{3}$ Anh Nguyen is with the Department of Computer Science, University of Liverpool, 
           {\tt\small anh.nguyen@liverpool.ac.uk}}           
    }

\begin{document}
\maketitle
\thispagestyle{empty}
\pagestyle{empty}

\begin{abstract}
This paper presents a new trajectory replanner for grasping irregular objects. Unlike conventional grasping tasks where the object's geometry is assumed simple, we aim to achieve a ``dynamic grasp'' of the irregular objects, which requires continuous adjustment during the grasping process. To effectively handle irregular objects, we propose a trajectory optimization framework that comprises two phases. Firstly, in a specified time limit of \SI{10}{\second}, initial offline trajectories are computed for a seamless motion from an initial configuration of the robot to grasp the object and deliver it to a pre-defined target location. 
Secondly, fast online trajectory optimization is implemented to update robot trajectories in real-time within \SI{100}{\milli\second}. This helps to mitigate pose estimation errors from the vision system. 
To account for model inaccuracies, disturbances, and other non-modeled effects, trajectory tracking controllers for both the robot and the gripper are implemented to execute the optimal trajectories from the proposed framework. The intensive experimental results effectively demonstrate the performance of our trajectory planning framework in both simulation and real-world scenarios. 
\end{abstract}

\section{Introduction}
\label{section: introduction}
\begin{figure} [!t]
    \centering
    \scalebox{0.75}{
    \def\svgwidth{1\columnwidth}
\begingroup%
  \makeatletter%
  \providecommand\color[2][]{%
    \errmessage{(Inkscape) Color is used for the text in Inkscape, but the package 'color.sty' is not loaded}%
    \renewcommand\color[2][]{}%
  }%
  \providecommand\transparent[1]{%
    \errmessage{(Inkscape) Transparency is used (non-zero) for the text in Inkscape, but the package 'transparent.sty' is not loaded}%
    \renewcommand\transparent[1]{}%
  }%
  \providecommand\rotatebox[2]{#2}%
  \newcommand*\fsize{\dimexpr\f@size pt\relax}%
  \newcommand*\lineheight[1]{\fontsize{\fsize}{#1\fsize}\selectfont}%
  \ifx\svgwidth\undefined%
    \setlength{\unitlength}{185.09999193bp}%
    \ifx\svgscale\undefined%
      \relax%
    \else%
      \setlength{\unitlength}{\unitlength * \real{\svgscale}}%
    \fi%
  \else%
    \setlength{\unitlength}{\svgwidth}%
  \fi%
  \global\let\svgwidth\undefined%
  \global\let\svgscale\undefined%
  \makeatother%
  \begin{picture}(1,1.36628865)%
    \lineheight{1}%
    \setlength\tabcolsep{0pt}%
    \put(0,0){\includegraphics[width=\unitlength,page=1]{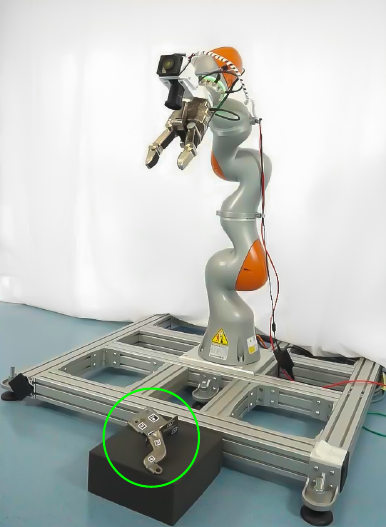}}%
    \put(0.01536279,0.61052237){\color[rgb]{0,0,0}\makebox(0,0)[lt]{\lineheight{1.25}\smash{\begin{tabular}[t]{l}\scalebox{0.8}{Grasping object}\end{tabular}}}}%
    \put(0,0){\includegraphics[width=\unitlength,page=2]{example_setup.pdf}}%
    \put(0.009626,0.87567755){\color[rgb]{0,0,0}\makebox(0,0)[lt]{\lineheight{1.25}\smash{\begin{tabular}[t]{l}\scalebox{0.8}{\schunk}\end{tabular}}}}%
    \put(0,0){\includegraphics[width=\unitlength,page=3]{example_setup.pdf}}%
    \put(0.01290531,1.23438676){\color[rgb]{0,0,0}\makebox(0,0)[lt]{\lineheight{1.25}\smash{\begin{tabular}[t]{l}\scalebox{0.8}{\camera}\end{tabular}}}}%
    \put(0,0){\includegraphics[width=\unitlength,page=4]{example_setup.pdf}}%
  \end{picture}%
\endgroup%

    }
    \caption{An example setup of grasping irregular objects.}%
    \vspace{-4ex}
    \label{fig: example setup captured}%
\end{figure}

Robots have become popular in the manufacturing industry because of the demand for increased automation and advances in computer processors. Besides, modular production setup allowing components to be changed has been trending in several industry sectors. Since robots are easily interchangeable for different purposes, they play a central role in rapidly adapting to new products and production processes. 
This study focuses on a proof-of-concept for a robot-assisted application. Specifically, grasping \textit{irregular objects} with complex geometries in producing engine casing parts (Fig. \ref{fig: example setup captured}). 
While typical robotic applications involve grasping objects from fixed positions~\cite{nguyen2016preparatory,vuong2024language}, this work investigates scenarios where the object's placement is random, which increases the complexity but improves the production line's flexibility. 
The research focuses on achieving \textit{flexibility} and \textit{dynamic grasping} capabilities where the robot remains in motion during the grasping process, which reduces application time and increases production efficiency. 

\subsection{Problem description}
To facilitate the seamless grasping task, our equipment comprises a \kuka robot, SDH2 gripper, \camera camera, and a grasping object, see Fig. \ref{fig: example setup captured}. The camera, mounted on the robot end-effector, follows an eye-in-hand configuration. Without loss of generality, we consider the grasping pose to be located at $\mathcal{O}_g$, which can be computed via a computer vision module at \SI{30}{\hertz} (e.g., AruCo marker and an extended Kalman filter). Note that since the object has an irregular shape and is not symmetrical, a small error caused by either the vision system or the trajectory planner can lead to unbalanced forces on the object's surface, causing internal dislocation of the object between two gripper fingers. Therefore, the \textit{replanning capability} of the proposed framework is critical. 
\subsection{Literature review}
\subsubsection {Optimization-based trajectory optimization}
Motion planning can be addressed through numerical optimization to determine a locally optimal trajectory, considering all dynamic constraints of the system \cite{betts1998survey,rao2009survey,vu2023machine}. 
Two notably successful algorithms for this purpose are CHOMP \cite{zucker2013chomp} and TrajOpt \cite{schulman2014motion}. In these algorithms, the trajectory is parameterized by its path progress or time. Then, the gradient-based methods are employed to identify the locally optimal trajectory. While these methods have proven successful in numerous applications \cite{zucker2013chomp}, their computation time remains too long for real-time implementation on a standard electronic control unit, as observed in studies such as \cite{iftikhar2019nonlinear, zhang2020optimization}. 
In authors' previous works \cite{minh1, minh2,vu2022fast}, a distinction is drawn between offline and online trajectory planning. Offline trajectory planning is conducted before the robot's movement to compute an optimal initial guess, whereas online trajectory planning occurs during execution. 
Recently, stochastic trajectory optimization (STO) methods, e.g., Via Point-STO \cite{jankowski2023vp}, Chance-Constrained Via Point-STO \cite{brudermuller2024cc}, have been utilized to optimize over a continuous trajectory space defined by via points. Constraints such as system limits are specified implicitly and handled by the trajectory representation. 
Although these mentioned approaches have been successfully implemented in different robotic systems, the online capability is still challenging for this grasping task. 
\subsubsection{Model predictive control-based trajectory optimization}
In recent years, there has been growing interest in extending trajectory optimization to online planning using receding horizons, encompassing gradient-based methods \cite{Schoels2020a} and sampling-based approaches \cite{Bhardwaj2022}. However, for manipulation tasks, simple point-to-point planning often falls short, necessitating the inclusion of additional constraints, such as pre-grasp points. Model Predictive Path Integral control \cite{gandhi2021robust, honda2024stein,williams2017model}, also called sampling-based MPC, has proven real-time performance on real robotic systems in challenging and dynamic environments. However,  these methods are typically limited to short-horizon problems \cite{jankowski2023vp}. 
Toussaint et al. \cite{Toussaint2022} introduced a sequence-of-constraints Model Predictive Control (MPC) approach to address task and motion planning (TAMP) in three stages. First, task planning generates waypoints; next, these waypoints are optimized in terms of timing to create a reference trajectory. Finally, MPC uses this reference to calculate collision-free paths over a short planning horizon. A global reference is essential for integrating waypoints into the MPC.

Inspired by the two phases mentioned above of trajectory planning and the MPC-based trajectory optimization, we first compute the offline trajectories for three phases, e.g., moving to the object, grasping the object, and moving to the target. Then, in an MPC fashion, a novel online trajectory planner is implemented to account for the new object pose update from the vision system.   
\subsection{Paper's contributions}
Our contributions are threefold. 
\begin{itemize}
    \item \emph{Firstly}, our method synchronizes offline trajectories for the robot and the gripper, considering the robot's dynamics, within a short computing time of $\SI{10}{\second}$, allowing dynamic grasping without disrupting the robot's motion. 
    \item \emph{Secondly}, our new online trajectory replanning method reactively adjusts the robot's and the gripper's trajectories within $\SI{100}{\milli\second}$ from the real-time updates of the object's pose. This responsiveness is crucial for precisely grasping objects with complex geometry, minimizing the impact of minor errors in position detection. 
    \item \emph{Thirdly}, experiments are demonstrated to show the generalization of the proposed framework. The proposed framework can be generalized to different grippers and grasping a moving object. Videos of experiments are found at \url{acin.tuwien.ac.at/39bb}
\end{itemize}

\section{Modeling}
 \graphicspath{{./}{graphics/}}
 \label{sec: modeling}
This section briefly presents the modeling of the robot and the \schunk (SDH2) gripper. 

\subsection{Modeling of the \kuka robot}
\begin{figure}
    \centering
    \scalebox{0.7}{
    \def\svgwidth{1\columnwidth}
    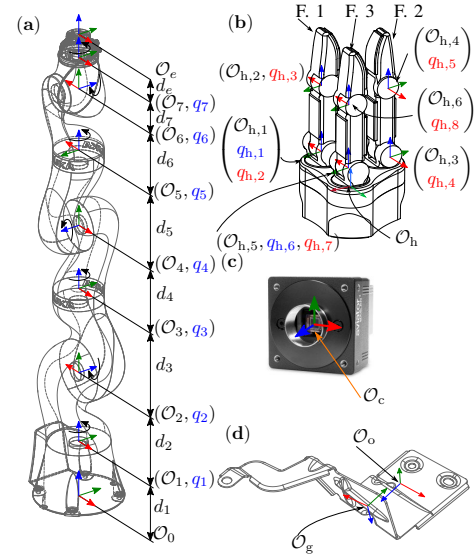
    }
    \caption{Schematic drawing of (a) the \kuka, (b) the SDH 2 hand, (c) the camera \camera, and (d) the object. The $x$-, $y$-, and $z$-axis of each coordinate frame are depicted by red, green, and blue arrows, respectively. In the tuple of the rotation angle
    and its corresponding frame $(\mathcal{O}_{\hsym,i},q_{\hsym,i})$, the illustrating color of the joint angle $q_{\hsym,i}$ matches the corresponding rotate axis.}%
    \label{fig: schematics}%
\end{figure}
The robot is modeled as a rigid-body system with the generalized coordinates $\mathbf{q}^\mathrm{T} = [q_1,q_2,\dots,q_7]$, see Fig. \ref{fig: schematics}(a), which are the rotation angles $q_i$ around the $z$-axes (blue arrows) of each coordinate frame $\mathcal{O}_i$, $i=1,\dots,7$. 
Considering system state $\mathbf{x}^{\mathrm{T}} = [\mathbf{q}^{\mathrm{T}},\dot{\mathbf{q}}^{\mathrm{T}}]$ and $\mathbf{v}$ is the vector of control inputs, the state-space form system dynamics is expressed as
\begin{equation}
    \dot{\mathbf{x}} = \mathbf{f}(\mathbf{x},\mathbf{v}) = [(\dot{\mathbf{q}})^\mathrm{T},\mathbf{v}^\mathrm{T}]^\mathrm{T}
    \label{eq: remaining dynamics state-space form}
\end{equation}

\subsection{Modeling of \schunk gripper}
\begin{figure}
    \centering
    \scalebox{0.7}
    {
    \def\svgwidth{1\columnwidth}
    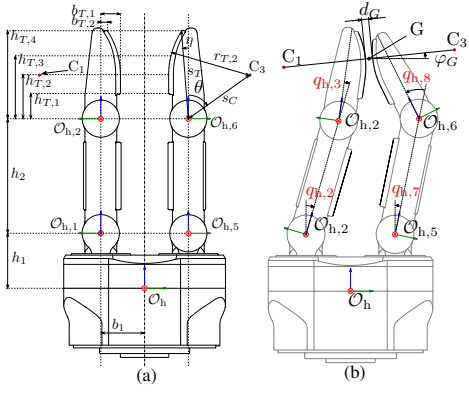
    }
    \caption{(a) Side view of the SDH2 without the finger 2 with $q_{\hsym,1}=-q_{\hsym,6}=-\pi/2$. (b) Side view of the gripping hand with the SDH2 state variables $\vec{q}_\hsym$ and the grasping state $\vec{q}_G$.}%
    \label{fig: SDH2 schematics}%
\end{figure}

The system state of the SDH2 consists of four DoF $\mathbf{q}_\hsym^\mathrm{T} = [q_{\hsym,2},q_{\hsym,3},q_{\hsym,7},q_{\hsym,8}]$, see Fig. \ref{fig: SDH2 schematics}(b). The grasping (contact) point $\mathrm{G}$ is chosen as the midpoint of the line between the points $\mathrm{C}_1$ and $\mathrm{C}_3$. 
These points are the centers of arcs formed by the tactile sensor surfaces of fingers 1 and 3, respectively. 
Note that the choice of this grasping point is crucial to achieve the steady state movement of the object after grasping. 
This is because the contact points at the object's surface will be perpendicular to the arcs of tactile sensor surfaces. Thereby, the contact forces are equally distributed to the object's surface. 
The forward kinematics formulation
\begin{equation}
    \vec{p}_{\hsym}^{\mathrm{G}}=[x_\mathrm{G},y_\mathrm{G},z_\mathrm{G}]^\mathrm{T} = \mathrm{SHD2\_FK}(\vec{q}_\hsym)
    \label{eq: SDH2 FK}
\end{equation}
describes the position of the grasping point $\mathrm{G}$ depending on the joint coordinates $\vec{q}_\hsym$.

\section{Trajectory optimization}
\label{chapter: trajectory optimizaton}
\begin{figure}[!t]
    \centering
    \scalebox{0.8}{
    \def\svgwidth{1\columnwidth}
    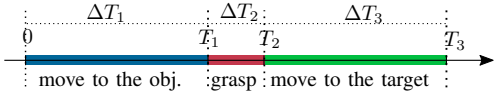
    }
    \caption{Timeline}%
    \vspace{-4ex}
    \label{fig: timeline}%
\end{figure}

This section presents the overall optimization framework, which consists of two stages. In the offline stage, we consider three trajectory optimization phases; see Fig \ref{fig: timeline}. 
In the first blue region of the timeline, the trajectory guides the robot from the initial configuration to the robot's target pose before the grasping process during the time $0\leq t \leq T_1$. 
Subsequently, the second trajectory phase $\Delta T_2$, discussed in Section \ref{ch:trajectories_robot_part2}, helps to move the robot from the robot's target pose before the grasping process to the target robot's pose after the grasping process during the time $T_1 < t \leq T_2$. Finally, Section \ref{ch:trajectories_robot_part3} describes the trajectory computation that brings the object to a user-defined target position during the time $T_2 < t \leq T_3$. 

\subsection{The offline trajectory optimization of the robot}
\label{section: offline opt}
In the $i$-th trajectory phase, $i\in{1,...,3}$, the trajectory that moves the robot from an initial configuration $\mathbf{x}_{i,S}$ to the target configuration $\mathbf{x}_{i,T_i}$ is formulated as an optimization by discretizing the trajectory $\bm{\xi}_{i}(t), t\in [0,\Delta T_i]$, with $N_i+1$ grid points and solving the resulting static optimization problem 
\begin{subequations}\label{eq:opt_prob_traj_1}
	\begin{align}
		\underset{\bm{\xi}_i}{\min}\quad\;\; &J\left(\bm{\xi}_i\right) = \Delta T_i + \Delta t_i \sum_{k=0}^{N_i} \transpose{\vec{v}_{1,k}}\vec{v}_{1,k}\\
		\label{eq:trapezoidal_method_traj1}
		\text{s.t.} \quad &\mathbf{x}_{i,k+1} - \mathbf{x}_{i,k} = \dfrac{1}{2}\Delta t_\hsym\begin{bmatrix}
                \dot{\mathbf{q}}_{i,k+1} + \dot{\mathbf{q}}_{i,k} \\
                \mathbf{v}_{i,k+1} + \mathbf{v}_{i,k}
                \end{bmatrix} 
                \\
            \label{eq: traj 1 init and end state}                
		&\vec{x}_{i,0} = \vec{x}_{i,S}, \quad \vec{x}_{i,N} = \vec{x}_{i,T_i}\\
            \label{eq: traj 1 state constraint}  
		&\underline{\vec{x}} \leq \vec{x}_{i,k} \leq \overline{\vec{x}},\\
            \label{eq: traj 1 control input}  
            &\underline{\bm{\tau}} - \underline{\mathbf{c}} \leq ({\mathbf{M}}(\mathbf{q}_{i,k})\mathbf{v}_{i,k} + {\mathbf{g}}(\mathbf{q}_{i,k})) \leq \overline{\bm{\tau}} - \overline{\mathbf{c}}\\
            & k = 0, \dots, N_i \nonumber
	\end{align}
\end{subequations}
for the optimal trajectory
\begin{equation}
\transpose{(\bm{\xi}_{i}^*)} = \left[\Delta T_i^*,\transpose{(\vec{x}_{i,0}^*)}, \dots, \transpose{(\vec{x}_{i,N_i}^*)}, \transpose{(\vec{v}_{i,0}^*)}, \dots, \transpose{(\vec{v}_{i,N_i}^*)}\right],
\label{eq: traj 1 optimal val}
\end{equation}
with the time step $\Delta t_i = \Delta T_i /N_i$. 
Note that the final time $\Delta T_i^*$ in (\ref{eq: traj 1 optimal val}) denotes the optimal duration of the trajectory from the initial state $\mathbf{x}_{i,S}$ to the target configuration $\mathbf{q}_{i,T_1}$. The system dynamics (\ref{eq: remaining dynamics state-space form}) is approximated by the trapezoidal rule in (\ref{eq:trapezoidal_method_traj1}). 
Additionally, $\underline{\mathbf{x}}$ and $\overline{\mathbf{x}}$ in (\ref{eq: traj 1 state constraint}) denote the symmetric lower and upper bounds of the state, respectively, and (\ref{eq: traj 1 control input}) considers the upper and lower torque limit $\overline{\bm{\tau}}$ and $\underline{\bm{\tau}}$. 
Note that the torque limits in (\ref{eq: traj 1 control input}) are an expensive inequality constraint.
Instead of fully neglecting the Coriolis matrix ${\mathbf{C}}(\mathbf{q},\dot{\mathbf{q}})$, which is often done in industrial applications \cite{binder1986distributed}, upper and lower bounds, $\overline{\mathbf{c}}$ and $\underline{\mathbf{c}}$, for the values of ${\mathbf{C}}(\mathbf{q},\dot{\mathbf{q}})$ are determined according to \cite{minh1}.
Although the influence of the Coriolis matrix on the overall system's dynamics is insignificant, it is still advantageous to consider these physical limits in the optimization problem (\ref{eq:opt_prob_traj_1}). 
Note that to create smooth transitions, the final state of a trajectory is chosen as the initial state of the subsequent trajectory. 
The following presents additional constraints to (\ref{eq:opt_prob_traj_1}) for each phase. 
\subsubsection{Phase $1$}
\label{ch:trajectories_robot_part1}
\textit{Trajectory of approaching the object from an initial configuration}

\begin{figure}[!t]
    \centering
    \scalebox{0.5}{
    \def\svgwidth{1\columnwidth}
    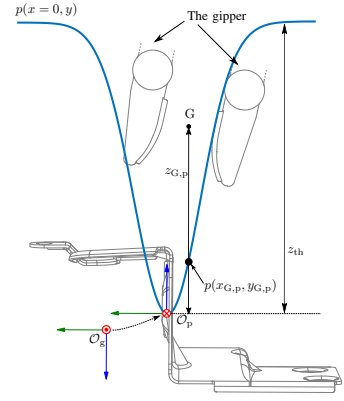
    }
    \caption{Side view of the grasping object and fingertips with the potential function $p\left(0,y\right)$.}%
    \vspace{-3ex}
    \label{fig:haerteteil_2_grip_sideview_gauss}%
\end{figure}

Since the object to be grasped has an unconventional shape, see Fig. \ref{fig: schematics}(d), the potential function is employed that restricts the motion of the gripper while approaching the object. 
In Fig. \ref{fig:haerteteil_2_grip_sideview_gauss}, a simple description of the potential function is illustrated where a slide view of a potential function is drawn in blue. 
Therein, the motion of the grasping point $\mathrm{G}$ of the gripper is restricted in the inverted Gaussian bell shape (illustrated in blue color) when the height of the point $\mathrm{G}$ is smaller than a threshold value $z_\mathrm{th}$. 
In the following, the potential function is described in detail. The potential function is formulated as an inverted Gaussian function in the form
\begin{equation}
p\left(x,y\right) = z_\mathrm{th}\bigg(1 -\exp\bigg(-\frac{1}{2}
\begin{bmatrix} x-\mu_x\\y-\mu_y \end{bmatrix}^\mathrm{T}
\bm{\Sigma}^{-1}
\begin{bmatrix} x-\mu_x\\y-\mu_y \end{bmatrix}
\bigg)
\bigg) \Comma
\label{eq:bidimensional_gauss} 
\end{equation}
where ${z}_\mathrm{th}$ is the threshold value, $\bm{\Sigma}$ is the covariance matrix, and $\bm{\mu}^\mathrm{T} = [\mu_x,\mu_y]$ is mean value of the Gaussian function. 
In Fig. \ref{fig:haerteteil_2_grip_sideview_gauss}, a slide view of the potential function $p(x=0,y)$ in the frame $\mathcal{O}_\psym$ is depicted with the variances are chosen to be $\sigma_x^2=\SI{5}{\milli\meter}$ and $\sigma_y^2=\SI{2}{\milli\meter}$. The mean value is $\bm{\mu}^\mathrm{T} = [0,0]$. 

Thereby, to guarantee that the grasping point is inside the region of the Gaussian bell shape created by the potential function, the following condition
\begin{equation}
    h(\mathbf{q}) = z_{\mathrm{G},p} - p(x_{\mathrm{G},p},y_{\mathrm{G},p}) \geq 0
\label{eq: gaussian constraint}
\end{equation}
must be added in (\ref{eq:opt_prob_traj_1}) for the phase $i=1$.

\subsubsection{Phase $2$}
\label{ch:trajectories_robot_part2}
\textit{Trajectory of the robot during grasping action of the gripper}

In addition to (\ref{eq:opt_prob_traj_1}) for $i=2$, to ensure the object is lifted from the ground at the end of this second trajectory phase, the $z$-axis velocity $\vec{v}_{0,T_2}^\hsym$ of the gripper w.r.t. the world frame $\mathcal{O}_0$ is larger than $0$, corresponding to
\begin{equation}
	\transpose{\vec{e}_z}\vec{v}_{0,T_2}^\hsym\geq 0,\quad 
        \mathrm{with} \quad \vec{v}_{0,T_2}^\hsym = \dfrac{\partial \mathbf{p}_{0}^\hsym}{\partial \mathbf{q}} \Biggr|_{\mathbf{q}=\mathbf{q}_{2,T_2}}\dot{\vec{q}}_{2,T_2} \FullStop
\label{eq: traj 2 condition vel}        
\end{equation}

\subsubsection{Phase 3}\label{ch:trajectories_robot_part3}
\textit{Trajectory of the movement from the grasping object to the target position}

In this phase (green region in Fig. \ref{fig: timeline}), the optimal trajectory $\bm{\xi}_3$ obtained from (\ref{eq:opt_prob_traj_1}) drives the robot from the final state $\mathbf{x}_{2,T_2}$ of phase 2 to a predefined target state $\mathbf{x}_{3, T_3}$. Note that the target state is a stationary point, i.e., $\mathbf{x}_{3, T_3} = [\mathbf{q}_{3,T_3},\mathbf{0}_3]$.

\subsection{Online trajectory replanner}
\label{sec: online update}

Due to the complexity of optimization problems, the optimization (\ref{eq:opt_prob_traj_1}) requires longer computation times ($\geq$\SI{3}{\second}) with dense grid points $N_i=100,\:i=\{1,2,3\}$. 
Therefore, the following procedure presents the online trajectory replanner to adapt the robot trajectories $\bm{\xi}_i^*,\: i = \{1,2,3\}$. 
To precisely grasp the object, the pose of the object $\hat{\mathbf{H}}_0^{o}$ is updated using the computer vision module at the rate $10$ Hz. 
Thereby, the new starting $\hat{\mathbf{x}}_{S_i}$ and target states $\hat{\mathbf{x}}_{T_i},\: i=\{1,2,3\}$ are computed accordingly. 
Here, the online trajectory replanner is used to minimize the deviation from the current optimal trajectory while satisfying the dynamic constraints of the system and adapting these new target states. 
Since the following procedures are identical for three trajectory phases $\bm{\xi}_i,\:i\in\{1,2,3\}$, the subscript $i$ is omitted for compact notations. 
The current optimal trajectory is expressed as
\begin{equation}
(\bm{\xi}^*)^\mathrm{T} = [\Delta T^*, (\mathbf{x}^*_0)^\mathrm{T}, \dots, (\mathbf{x}^*_N)^\mathrm{T},(\mathbf{v}^*_0)^\mathrm{T}, \dots, (\mathbf{v}^*_N)^\mathrm{T}]\FullStop
\label{eq: refine}
\end{equation}
Note that this optimal trajectory satisfies the dynamical constraints approximated by the trapezoidal rule in the form
\begin{equation}
    \mathbf{x}_{k+1}^* = \mathbf{x}_k^* + \dfrac{\Delta T}{2N}(\mathbf{f}_k^*+\mathbf{f}_{k+1}^*)\FullStop
    \label{eq: trapezoidal rule optimal traj}
\end{equation}
For the online trajector replanner, small deviations 
\begin{equation*}
\delta\bm{\xi}^\mathrm{T} = [\delta T,(\delta\mathbf{x}_0)^\mathrm{T}, . ..., (\delta\mathbf{x}_N)^\mathrm{T},(\delta\mathbf{v}_0)^\mathrm{T}, ...,(\delta \mathbf{v}_N)^\mathrm{T}]
\end{equation*}
need to be taken into account to compute the new trajectory connecting the current starting state $\hat{\mathbf{x}}_S$ with the updated target state $\hat{\mathbf{x}}_T$. 
The discrete-time system dynamics (\ref{eq: remaining dynamics state-space form}) w.r.t. the interpolated optimal state $\bm{\xi}^*$ reads as
\begin{equation}
\begin{aligned}
\mathbf{x}_{k+1} & = \mathbf{x}_k + \dfrac{\Delta T^* + \delta T}{2N} \bigg(\mathbf{f}^*_k + \bm{\Gamma}_{k}^{\mathbf{x}}\delta \mathbf{x}_k +\bm{\Gamma}_{k}^{\mathbf{v}}\delta \mathbf{v}_k \\
&+ \mathbf{f}^*_{k+1}+ \bm{\Gamma}_{k+1}^{\mathbf{x}}\delta \mathbf{x}_{k+1} + \bm{\Gamma}_{k+1}^{\mathbf{v}}\delta \mathbf{v}_{k+1} \bigg), 
\end{aligned}
\label{eq: linearization sd}
\end{equation}
with $\delta\mathbf{x}_k = \mathbf{x}_k - \mathbf{x}^*_{k}$, $\delta\mathbf{v}_k = \mathbf{v}_k - \mathbf{v}^*_{k}$, $\delta t = \Delta T - \Delta T^*$, and
\begin{equation*}
\bm{\Gamma}_{k}^{\mathbf{x}} = \dfrac{\partial \mathbf{f}}{\partial \mathbf{x}}\Bigr|_{\substack{\mathbf{x}_k^*,\mathbf{v}_k^*}}, \:\:\: \bm{\Gamma}_{k}^{\mathbf{v}} = \dfrac{\partial \mathbf{f}}{\partial \mathbf{v}}\Bigr|_{\substack{\mathbf{x}_k^*,\mathbf{v}_k^*}}
\end{equation*}
for  $k=0,...,N-1$. 
Subtracting (\ref{eq: trapezoidal rule optimal traj}) from (\ref{eq: linearization sd}) and neglecting terms containing a product of deviation variables, we have 
\begin{equation}
\begin{aligned}
\delta \mathbf{x}_{k+1}  = \delta \mathbf{x}_k &+ \dfrac{\Delta T^*}{2N}\bigg(\bm{\Gamma}_{k}^{\mathbf{x}}\delta \mathbf{x}_k +\bm{\Gamma}_{k}^{\mathbf{v}}\delta \mathbf{v}_k +  \bm{\Gamma}_{k+1}^{\mathbf{x}}\delta \mathbf{x}_{k+1} \\
&+ \bm{\Gamma}_{k+1}^{\mathbf{v}}\delta \mathbf{v}_{k+1} \bigg) + \dfrac{\delta T}{\Delta T^*} (\mathbf{x}^*_{k+1} - \mathbf{x}^*_{k}).
\end{aligned}
\label{eq: linearization new}
\end{equation}
In a more compact form, (\ref{eq: linearization new}) is rewritten as
\begin{equation}
\mathbf{C}_{k+1} \mathbf{s}_{k+1} = \mathbf{A}_k \mathbf{s}_k,
\label{eq: mpc form}
\end{equation}
where 
\begin{equation*}
\begin{aligned}
\mathbf{C}_{k+1} &= \begin{bmatrix}
\mathbf{I} - \dfrac{h^*}{2}\bm{\Gamma}_{k+1}^{\mathbf{x}} &
- \dfrac{h^*}{2} \bm{\Gamma}_{k+1}^{\mathbf{v}} & 0 \\
\mathbf{0} & \mathbf{0} & 1 
\end{bmatrix},
\\
\mathbf{A}_{k} &= \begin{bmatrix}
\mathbf{I} + \dfrac{h^*}{2}\bm{\Gamma}_{k}^{\mathbf{x}} & 
\dfrac{h^*}{2} \bm{\Gamma}_{k}^{\mathbf{v}} &
\dfrac{\mathbf{x}^*_{k+1} - \mathbf{x}^*_{k}}{\Delta T^*}
\\ 
\mathbf{0} & \mathbf{0} & 1 
\end{bmatrix},
\\
\mathbf{s}_k &= \begin{bmatrix}
\delta \mathbf{x}_k \\
\delta \mathbf{v}_k \\
\delta T_{k}
\end{bmatrix},
\end{aligned}
\end{equation*}
and $h^* = \dfrac{\Delta T^*}{N}$. For simplicity, only one variable for the final time $\delta T$ in (\ref{eq: linearization sd}) was introduced instead of $\delta T_{k}, k=0,\dots,N-1$. Thus, $\delta T_{k+1} = \delta T_{k}$ was used in (\ref{eq: mpc form}). The deviation vector $\bm{\xi}_k$ is obtained as the solution of a linear constrained quadratic program (LCQP) of the form
\begin{subequations}\label{eq: QP}
\begin{align}
\label{eq: QP a}
 \min_{\mathbf{s}_k}  \:\:\: &\dfrac{1}{2}\sum_{k=1}^{N-1}\mathbf{s}_{k}^\mathrm{T}\mathbf{Q}_k \mathbf{s}_k \\
\label{eq: QP b}
\text{s.t.} \:\: &\mathbf{C}_{k+1} \mathbf{s}_{k+1} = \mathbf{A}_k \mathbf{s}_k,\quad k = 0,...,N-1\\
\label{eq: QP c}
& \underline{\mathbf{s}_k} \leq \mathbf{s}_k  \leq \overline{\mathbf{s}_k}, \quad k=0,...,N 
\end{align}
\end{subequations}
%
and the positive definite weighting matrix
\begin{equation}
\mathbf{Q}_k = \text{diag}(\mathbf{Q}_{\mathbf{x}_k},\mathbf{Q}_{\mathbf{v}_k},Q_{\Delta T}).
\label{eq: weighting matrix}
\end{equation}
With the choice of $Q_{t_F}>0$ and the submatrices $\mathbf{Q}_{\mathbf{x}_k}$ and $\mathbf{Q}_{\mathbf{v}_k}$, the deviation of the online trajectory from the trajectory (\ref{eq: refine}) can be weighted explicitly in the objective function (\ref{eq: QP a}) w.r.t. the traversal time $\Delta T$, the state $\mathbf{x}_k$, and the control input $\mathbf{v}_k$, respectively. 
The inequality condition (\ref{eq: QP c}) allows the admissible tolerances of the online trajectory, where 
\begin{equation*}
\begin{aligned}
\underline{\mathbf{s}_k}^\mathrm{T} &= [\underline{\mathbf{x}_k}^\mathrm{T}-(\mathbf{x}_k^*)^\mathrm{T},\underline{\mathbf{v}_k}^\mathrm{T}-(\mathbf{v}_k^*)^\mathrm{T},\underline{\delta T}]\Comma \\
\overline{\mathbf{s}_k}^\mathrm{T} &= [\overline{\mathbf{x}_k}^\mathrm{T}-(\mathbf{x}_k^*)^\mathrm{T},\overline{\mathbf{v}_k}^\mathrm{T}-(\mathbf{v}_k^*)^\mathrm{T},\overline{\delta T}]\Comma
\end{aligned}
\end{equation*}
for $k=1,\dots,N-1$, and $\overline{\delta T}$ and $\underline{\delta T}$  is a sufficiently large upper and lower bound for $\delta t_F$, respectively. Additionally, the equality constraints must be taken into account by 
\begin{equation*}
\begin{aligned}
\underline{\mathbf{s}_0}^\mathrm{T} &= [\delta \mathbf{x}_0^\mathrm{T},\delta\mathbf{v}_0^\mathrm{T},\underline{\delta T}], \:\:
\overline{\mathbf{s}_0}^\mathrm{T} = [\delta \mathbf{x}_0^\mathrm{T}, \delta\mathbf{v}_0^\mathrm{T},\overline{\delta T}],
\\
\underline{\mathbf{s}_N}^\mathrm{T} &= [\delta \mathbf{x}_N^\mathrm{T},\delta\mathbf{v}_N^\mathrm{T},\underline{\delta T}],\:\: \overline{\mathbf{s}_N}^\mathrm{T} = [\delta \mathbf{x}_N^\mathrm{T},\delta \mathbf{v}_N^\mathrm{T},\overline{\delta T}]\FullStop
\end{aligned}
\end{equation*}
Finally, for each trajectory phase, the optimal trajectory of the online trajectory replanner $\bm{\xi}^{*}$ reads as
\begin{equation}
\bm{\xi}^{*} \leftarrow \bm{\xi}^{*} +  \delta \bm{\xi}^*\Comma
\label{eq: onl update}
\end{equation}
where $\delta \bm{\xi}^*$ results from the solution of (\ref{eq: QP}) in the form
\begin{equation*}
(\delta \bm{\xi}^*)^\mathrm{T} = [\delta T^*,(\delta \mathbf{x}_0^*)^\mathrm{T},...,(\delta \mathbf{x}_N^*)^\mathrm{T},(\delta \mathbf{v}_0^*)^\mathrm{T},...,(\delta \mathbf{v}_N^*)^\mathrm{T}]\FullStop
\end{equation*} 

\section{Results}
\label{section: results}
The experimental setup is illustrated in Fig. \ref{fig: example setup captured}, driven by the 4.0 GHz Intel Core i7-10700K PC with 32 GB RAM. Three network interface controllers (NIC) are used for communication with the robot, the SDH2 gripper, and the camera via EtherCAT, RT-Ethernet, and Ethernet, respectively.  
Since the joint velocities of the complete system cannot be measured directly, differential filters with a time constant of $T_1 =$ \SI{12}{\milli\second} are used for each joint. 
The sampling time for the controller is $T_s=125$ \SI{}{\micro\second}. The offline trajectory optimization (\ref{eq:opt_prob_traj_1}) and the online replanner (\ref{eq: QP}) are solved using the \textit{Interior Point Optimizer} (IPOPT) \cite{ipopt} with the linear solver MA57 \cite{hsl,ma57}. In these optimization problems, the numbers of grid points $N_i, \: i\in\{1,2,3\}$ are set to $100$, and the average computing times are listed in Table \ref{tab: average comp}. 

\subsection{Simulation results}
\noindent\textbf{Statistical results.} We report the average computing times are taken on 988 successful trials over the total 1000 trials of random uniformly distributed object poses 
$(\hat{\vec{p}}_0^\osym)^\mathrm{T}=\left[p_{0,x}^\osym,p_{0,y}^\osym, p_{0,z}^\osym\right]\in\rbraces{\underline{\vec{p}_0^\osym},\overline{\vec{p}_0^\osym}}$ with $(\underline{\vec{p}_0^\osym})^\mathrm{T}=\rbraces{\SI{0,6}{\meter},\SI{-0,1}{\meter},\SI{0,1}{\meter}}$ and $(\overline{\vec{p}_0^\osym})^\mathrm{T}=\rbraces{\SI{0,7}{\meter},\SI{0,1}{\meter},\SI{0,1}{\meter}}$. 
In $12$ failed trials, the optimization problems reach the iteration limits set to $100$. In simulations, the proposed framework shows excellent results with a success rate of approximately $98.8\%$. 
In Tab. \ref{tab: average comp}, the computation times of the offline phases do not exceed \SI{10}{\second}.

\begin{table}[h]
\vspace{-2ex}
    \captionsetup{width=0.5\textwidth}
    \caption{Average computation time comparison.} 
    \vspace{-2ex}
    \label{tab: COMAU MC}
    \begin{center}
        \begin{tabular}{c c c}
            \textbf{Phase} & \textbf{Ours} & \textbf{VP-STO}~\cite{jankowski2023vp}\\
            \hline
             phase 1 opt. traj. (\ref{eq:opt_prob_traj_1}) & $2.7 \pm 0.4$ (\SI{}{\second})& $7.2 \pm 1.6$ (\SI{}{\second}) \\
             phase 2 opt. traj. (\ref{eq:opt_prob_traj_1}) & $1.2 \pm 0.1$ (\SI{}{\second})& $9.8 \pm 3.7$ (\SI{}{\second}) \\
             phase 3 opt. traj. (\ref{eq:opt_prob_traj_1}) & $4.3 \pm 0.9$ (\SI{}{\second})& $11.5 \pm 2.3$ (\SI{}{\second}) \\
             online opt. traj. (\ref{eq: QP}) & $53.5 \pm 7.9$ (\SI{}{\milli\second})& NA \\
             \hline
        \end{tabular}
    \end{center}
\label{tab: average comp}  
\vspace{-4ex}
\end{table}

\noindent\textbf{Comparison.}
In addition, we utilize the via-point stochastic trajectory optimization (VP-STO) \cite{jankowski2023vp} to solve the offline trajectory planning phases. Although the success rate of VP-STO is $100\%$, our computational speed for offline trajectory planning outperforms VP-STO significantly. 
Note that the average computation time of the online replanner is below $\SI{100}{\milli\second}$, significantly faster than the computing time of the offline trajectory planning. 
\subsection{Real robot results} 
\noindent\textbf{Static scenarios.}
The initial and the target configuration of the robot are chosen as 
\begin{equation}\label{eq:init_pos}
    \transpose{\vec{q}_T}=\rbraces{\SI{55}{\degree},\SI{39.5}{\degree},\SI{-6.6}{\degree},\SI{-22}{\degree},\SI{-36.3}{\degree},\SI{110.6}{\degree},\SI{134}{\degree}}.
\end{equation}
and
\begin{equation}
	\transpose{\vec{q}_T}=\rbraces{\SI{0}{\degree},\SI{0}{\degree},\SI{0}{\degree},\SI{-40}{\degree},\SI{0}{\degree},\SI{100}{\degree},\SI{90}{\degree}}.
\end{equation}

\begin{figure} [t]
    \centering
    \scalebox{0.8}
    {
    \def\svgwidth{1\columnwidth}
    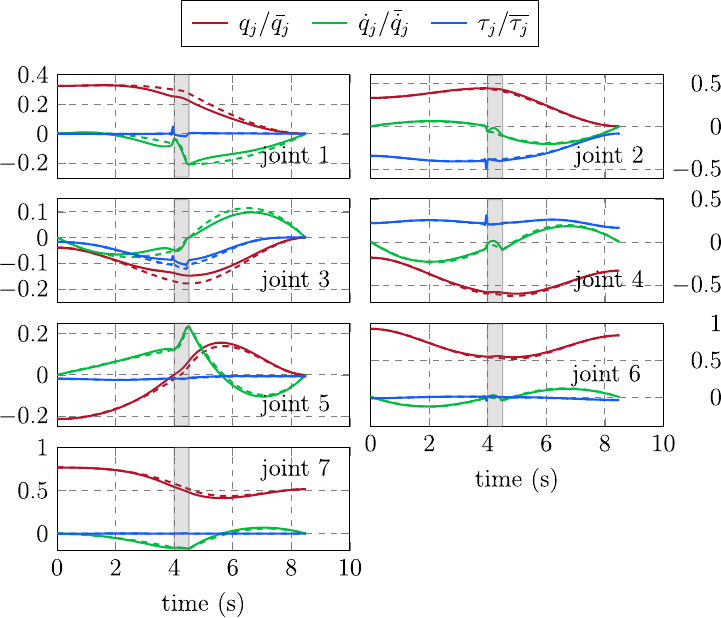
    }
    \vspace{2ex}
    \caption{Time evolution of the scaled offline and the online trajectory optimization with the initial robot configuration (\ref{eq:init_pos}). The dashed lines are the scaled offline trajectories $\bm{\xi}_i,i\in\{1,2,3\}$ obtained from (\ref{eq:opt_prob_traj_1}). The solid lines represent the scaled online trajectories obtained from (\ref{eq: onl update}). The grey areas indicate the second trajectory phase when the robot grasps the object. 
    The subscript $j$ indicates the $j^{th}$ joint of the robot. 
    }%
    \label{fig: time evolution experiment 1}%
\end{figure}

\begin{figure}[t]
    \centering
    \scalebox{0.8}
    {
   \def\svgwidth{1\columnwidth}  
   \includegraphics[width=\columnwidth,height=0.88\linewidth]{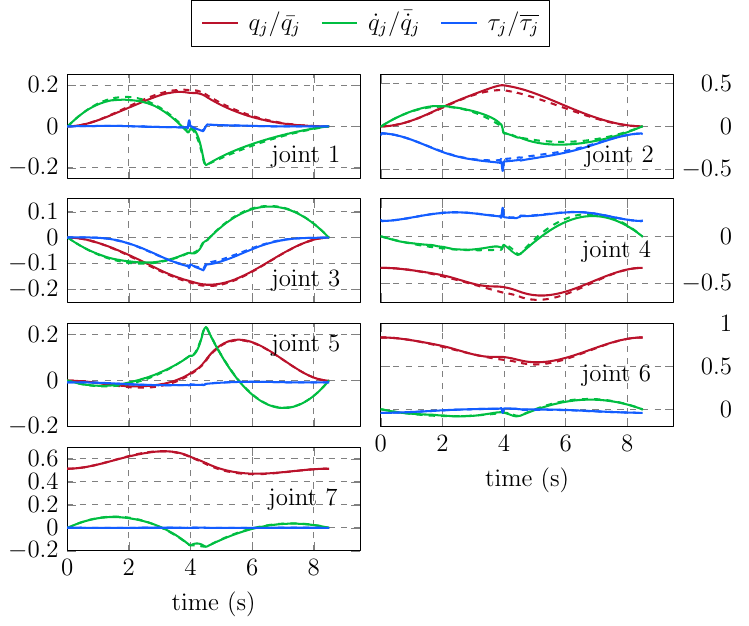}
   }
   \vspace{2ex}
   \caption{Time evolution of the experiment on the comparison of the proposed algorithm with and without the online replanning in (\ref{eq: QP}). The dashed lines illustrate the computed trajectory without the online replanner. The solid lines depict the computed trajectory from (\ref{eq: onl update}).}
   \vspace{-3ex}
    \label{fig: 3D experiment comparison time evolution}%
\end{figure}

The object is placed in the camera's field of view on the ground. With a given object pose from the computer vision system, the three-phase trajectory of the robot, illustrated by dashed lines in Fig. \ref{fig: time evolution experiment 1}, is computed offline using (\ref{eq:opt_prob_traj_1}). During the robot's motion, the online trajectories, depicted by solid lines, are updated periodically. In Fig. \ref{fig: time evolution experiment 1}, the grey areas indicate the second phase, i.e., the grasping process. It is worth noting that three parts of the robot's trajectory are smoothly connected. 
Since the trajectories in Fig. \ref{fig: time evolution experiment 1} do not surpass the $\pm 1$ horizontal line, hence, all the state and input constraints according to (\ref{eq: traj 1 state constraint}) and (\ref{eq: QP c}) are respected.

\begin{figure}[!t]
     \centering
     \scalebox{0.8}
     {
     \def\svgwidth{1\columnwidth}
     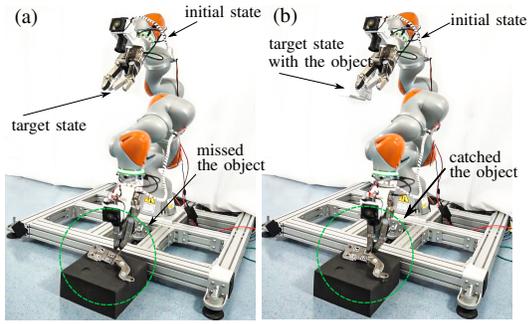
     }
     \caption{Snapshots of experiment on the comparison of the proposed trajectory optimization framework with (a) and without (b) the online replanning.}
     \vspace{-1ex}
     \label{fig: 3D experiment comparison snapshots}%
\end{figure}
\noindent\textbf{Comparison between with and without the online replanner.}
To further validate the effectiveness of the online replanner, a comparison with only the offline solution is conducted. To this end, two trials in which the initial and the target configuration of the robot are both chosen as 
\begin{equation}\label{eq:comparison init and target}
\transpose{\vec{q}_S}=\transpose{\vec{q}_T}=\rbraces{\SI{0}{\degree},\SI{0}{\degree},\SI{0}{\degree},\SI{-40}{\degree},\SI{0}{\degree},\SI{100}{\degree},\SI{90}{\degree}} \Comma
\end{equation}
and the object is randomly positioned in the camera's field of view, as shown in Fig. \ref{fig: 3D experiment comparison snapshots}. Only the offline trajectories computed from (\ref{eq:opt_prob_traj_1}) are sent to the robot controller in the first trial. On the other hand, in the second trial, the online replanner (\ref{eq: QP}) periodically updates the trajectory according to the object pose's update. The scaled trajectories of the first and second trials are illustrated by the dashed lines and the solid lines in Fig. \ref{fig: 3D experiment comparison time evolution}. 
Therein, during $0\leq t \leq 3$, the robot mainly follows the offline optimal trajectories. Once the robot is close to the object and a more precise object pose is acquired, minor deviations occur at $3 < t \leq 4$. Although the deviations are minor, the online trajectory replanner helps the robot to successfully grasp the object while only executing the offline trajectory failed to grasp the object, illustrated in Fig. \ref{fig: 3D experiment comparison snapshots} (a) and (b), respectively. 

\begin{table}[!t]
    \captionsetup{width=0.5\textwidth}
    \caption{Results of grasping moving objects over 20 trials.} 
    \label{tab: COMAU MC}
    \begin{center}
        \begin{tabular}{c c c}
             & \textbf{Ours} & \textbf{RRTConnect}~\cite{Coleman2014}\\
            \hline
             success rate (\%) & 95 & 75 \\
             comp. violation (num. of trials) & 0 & 2 \\
             missed object (num. of trials) & 1 & 3 \\
             \hline
        \label{tab: average MC}    
        \end{tabular}
    \end{center}
    \vspace{-5ex}
\end{table}

\begin{figure}[!t]
     \centering
     \scalebox{0.96}
     {
     \def\svgwidth{0.8\columnwidth}
     \input{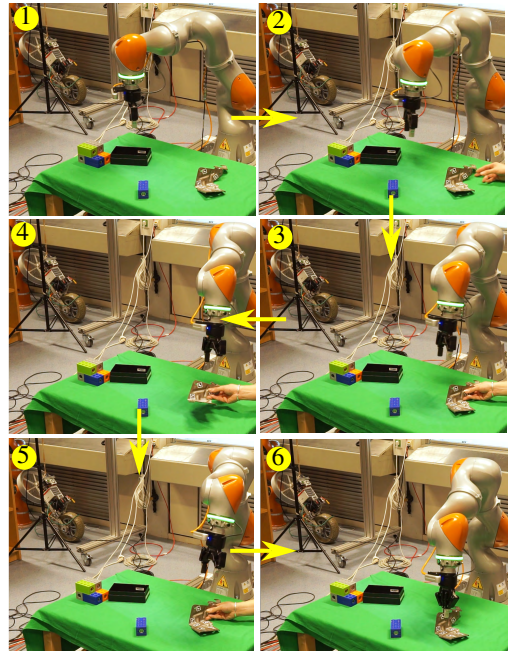}
     }
     \caption{A demonstration of grasping the moving object.}
     \vspace{-3ex}
     \label{fig: grasping moving object}%
\end{figure}

\noindent\textbf{Generalization to dynamic scenarios.} The proposed framework is generalized to a \textit{different gripper} (Robotiq 2F85) and can grasp a dynamically moving object. In Fig. \ref{fig: grasping moving object}, overlay images of an experiment are illustrated. When the robot nearly reaches the object, the object is suddenly relocated (see subfigures 2,3 in Figure \ref{fig: grasping moving object}). The online replanner can still adjust the offline trajectories and precisely grasp the object (see subfigure 6 in Fig. \ref{fig: grasping moving object}). We further run a small-scale Monte Carlo simulation of 20 trials to compare the performance of our proposed framework with RRTConnect~\cite{Kuffner2000} implemented in MoveIt~\cite{Coleman2014}. Statistical results are presented in Tab. \ref{tab: average MC}. 
Note that the Time-Optimal Trajectory Generation (TOTG) algorithm~\cite{Kunz2012} is applied to obtain the time parametrization for the RRTConnect algorithm. We chose RRTConnect over another variant of the RRT algorithm due to the computation speed, which is fast enough for this moving object scenario. The update frequency for the trajectory replanner is \SI{100}{\milli\second}. With our proposed approach, there is a $1$ failed trial when the robot misses grasping the object, i.e., 95\% success rate. On the other hand, the RRTConnect planner violated the time limit in 2 trials and failed to grasp the moving object in another three trials. 
Video of the experiments is provided at \url{acin.tuwien.ac.at/39bb} 

\section{Discussion}
\label{section: conclusions}
This paper presents the two-stage trajectory optimization framework for grasping irregular objects with complex geometry using a novel online trajectory replanning. To account for the object's pose estimation errors from a computer vision module, a novel online update of the robot trajectory is implemented to react to the change of the object's pose during the grasping motion. This application has been verified and tested in several experiments and outperforms the current state-of-the-art trajectory planning method in computational speed. 
The proposed framework is also generalized with different grippers and in moving object scenarios. With this extreme scenario, the proposed framework can successfully grasp the object in 19 over 20 trials. 
The main limitation of the proposed method is that it relies on offline trajectories requiring $\SI{10}{\second}$ for computation. Therefore, we are currently extending this two-stage approach to the online version, where we consider the shorter planning horizon for the offline phase and the incremental update when approaching the goal. Another direction is to compute and learn motion primitives for the offline planning phase via generative AI techniques, e.g., diffusion models, and variant autoencoder. As a result, this method could be used in an interactive scenario like language-driven human-robot collaboration tasks. 






\bibliography{RAL_grasping.bib}             

\begin{thebibliography}{10}
\providecommand{\url}[1]{#1}
\csname url@rmstyle\endcsname
\providecommand{\newblock}{\relax}
\providecommand{\bibinfo}[2]{#2}
\providecommand\BIBentrySTDinterwordspacing{\spaceskip=0pt\relax}
\providecommand\BIBentryALTinterwordstretchfactor{4}
\providecommand\BIBentryALTinterwordspacing{\spaceskip=\fontdimen2\font plus
\BIBentryALTinterwordstretchfactor\fontdimen3\font minus \fontdimen4\font\relax}
\providecommand\BIBforeignlanguage[2]{{%
\expandafter\ifx\csname l@#1\endcsname\relax
\typeout{** WARNING: IEEEtran.bst: No hyphenation pattern has been}%
\typeout{** loaded for the language `#1'. Using the pattern for}%
\typeout{** the default language instead.}%
\else
\language=\csname l@#1\endcsname
\fi
#2}}

\bibitem{nguyen2016preparatory}
A.~Nguyen, D.~Kanoulas, D.~G. Caldwell, and N.~G. Tsagarakis, ``Preparatory object reorientation for task-oriented grasping,'' in \emph{2016 IEEE/RSJ International Conference on Intelligent Robots and Systems (IROS)}.\hskip 1em plus 0.5em minus 0.4em\relax IEEE, 2016, pp. 893--899.

\bibitem{vuong2024language}
A.~D. Vuong, M.~N. Vu, B.~Huang, N.~Nguyen, H.~Le, T.~Vo, and A.~Nguyen, ``Language-driven grasp detection,'' in \emph{Proceedings of the IEEE/CVF Conference on Computer Vision and Pattern Recognition}, 2024, pp. 17\,902--17\,912.

\bibitem{betts1998survey}
J.~T. Betts, ``Survey of numerical methods for trajectory optimization,'' \emph{Journal of Guidance, Control, and Dynamics}, vol.~21, no.~2, pp. 193--207, 1998.

\bibitem{rao2009survey}
A.~V. Rao, ``A survey of numerical methods for optimal control,'' \emph{Advances in the Astronautical Sciences}, vol. 135, no.~1, pp. 497--528, 2009.

\bibitem{vu2023machine}
M.~N. Vu, F.~Beck, M.~Schwegel, C.~Hartl-Nesic, A.~Nguyen, and A.~Kugi, ``Machine learning-based framework for optimally solving the analytical inverse kinematics for redundant manipulators,'' \emph{Mechatronics}, vol.~91, p. 102970, 2023.

\bibitem{zucker2013chomp}
M.~Zucker, N.~Ratliff, A.~D. Dragan, M.~Pivtoraiko, M.~Klingensmith, C.~M. Dellin, J.~A. Bagnell, and S.~S. Srinivasa, ``{CHOMP}: Covariant hamiltonian optimization for motion planning,'' \emph{The International Journal of Robotics Research}, vol.~32, no. 9-10, pp. 1164--1193, 2013.

\bibitem{schulman2014motion}
J.~Schulman, Y.~Duan, J.~Ho, A.~Lee, I.~Awwal, H.~Bradlow, J.~Pan, S.~Patil, K.~Goldberg, and P.~Abbeel, ``Motion planning with sequential convex optimization and convex collision checking,'' \emph{The International Journal of Robotics Research}, vol.~33, no.~9, pp. 1251--1270, 2014.

\bibitem{iftikhar2019nonlinear}
S.~Iftikhar, O.~J. Faqir, and E.~C. Kemgan, ``Nonlinear model predictive control of an overhead laboratory-scale gantry crane with obstacle avoidance,'' \emph{Proceedings of the Conference on Control Technology and Applications (CCTA)}, pp. 382--387, 2019.

\bibitem{zhang2020optimization}
X.~Zhang, A.~Liniger, and F.~Borrelli, ``Optimization-based collision avoidance,'' \emph{IEEE Transactions on Control Systems Technology}, vol.~29, no.~3, pp. 972--983, 2020.

\bibitem{minh1}
M.~N. Vu, C.~Hartl-Nesic, and A.~Kugi, ``Fast swing-up trajectory optimization for a spherical pendulum on a 7-dof collaborative robot,'' in \emph{Proceedings of the International Conference on Robotics and Automation}, 2021, pp. 10\,114--10\,120.

\bibitem{minh2}
M.~Vu, P.~Zips, A.~Lobe, F.~Beck, W.~Kemmetm{\"u}ller, and A.~Kugi, ``Fast motion planning for a laboratory 3d gantry crane in the presence of obstacles,'' \emph{IFAC-PapersOnLine}, vol.~53, no.~2, pp. 9508--9514, 2020.

\bibitem{vu2022fast}
M.~N. Vu, A.~Lobe, F.~Beck, T.~Weingartshofer, C.~Hartl-Nesic, and A.~Kugi, ``Fast trajectory planning and control of a lab-scale 3d gantry crane for a moving target in an environment with obstacles,'' \emph{Control Engineering Practice}, vol. 126, p. 105255, 2022.

\bibitem{jankowski2023vp}
J.~Jankowski, L.~Bruderm{\"u}ller, N.~Hawes, and S.~Calinon, ``Vp-sto: Via-point-based stochastic trajectory optimization for reactive robot behavior,'' in \emph{2023 IEEE International Conference on Robotics and Automation (ICRA)}.\hskip 1em plus 0.5em minus 0.4em\relax IEEE, 2023, pp. 10\,125--10\,131.

\bibitem{brudermuller2024cc}
L.~Bruderm{\"u}ller, G.~Berger, J.~Jankowski, R.~Bhattacharyya, and N.~Hawes, ``Cc-vpsto: Chance-constrained via-point-based stochastic trajectory optimisation for safe and efficient online robot motion planning,'' \emph{arXiv preprint arXiv:2402.01370}, 2024.

\bibitem{Schoels2020a}
T.~Schoels, P.~Rutquist, L.~Palmieri, A.~Zanelli, K.~O. Arras, and M.~Diehl, ``{CIAO⁎}: {MPC}-based safe motion planning in predictable dynamic environments,'' \emph{IFAC-PapersOnLine}, vol.~53, no.~2, pp. 6555--6562, 2020.

\bibitem{Bhardwaj2022}
M.~Bhardwaj, B.~Sundaralingam, A.~Mousavian, N.~D. Ratliff, D.~Fox, F.~Ramos, and B.~Boots, ``Storm: An integrated framework for fast joint-space model-predictive control for reactive manipulation,'' in \emph{Proceedings of the Conference on Robot Learning (CoRL)}, vol. 164, 2022, pp. 750--759.

\bibitem{gandhi2021robust}
M.~S. Gandhi, B.~Vlahov, J.~Gibson, G.~Williams, and E.~A. Theodorou, ``Robust model predictive path integral control: Analysis and performance guarantees,'' \emph{IEEE Robotics and Automation Letters}, vol.~6, no.~2, pp. 1423--1430, 2021.

\bibitem{honda2024stein}
K.~Honda, N.~Akai, K.~Suzuki, M.~Aoki, H.~Hosogaya, H.~Okuda, and T.~Suzuki, ``Stein variational guided model predictive path integral control: Proposal and experiments with fast maneuvering vehicles,'' in \emph{2024 IEEE International Conference on Robotics and Automation (ICRA)}.\hskip 1em plus 0.5em minus 0.4em\relax IEEE, 2024, pp. 7020--7026.

\bibitem{williams2017model}
G.~Williams, A.~Aldrich, and E.~A. Theodorou, ``Model predictive path integral control: From theory to parallel computation,'' \emph{Journal of Guidance, Control, and Dynamics}, vol.~40, no.~2, pp. 344--357, 2017.

\bibitem{Toussaint2022}
M.~Toussaint, J.~Harris, J.-S. Ha, D.~Driess, and W.~Hönig, ``Sequence-of-constraints {MPC}: Reactive timing-optimal control of sequential manipulation,'' in \emph{Proceedings of the IEEE/RSJ International Conference on Intelligent Robots and Systems}, 2022, pp. 13\,753--13\,760.

\bibitem{binder1986distributed}
E.~E. Binder and J.~H. Herzog, ``Distributed computer architecture and fast parallel algorithms in real-time robot control,'' \emph{IEEE Transactions on Systems, Man, and Cybernetics}, vol.~16, no.~4, pp. 543--549, 1986.

\bibitem{ipopt}
A.~W\"{a}chter and L.~T. Biegler, ``On the implementation of an interior-point filter line-search algorithm for large-scale nonlinear programming,'' \emph{Math. Program.}, vol. 106, no.~1, pp. 25--57, 2006.

\bibitem{hsl}
\BIBentryALTinterwordspacing
HSL. (2018) A collection of fortran codes for large scale scientific computation. [Online]. Available: \url{http://www.hsl.rl.ac.uk/}
\BIBentrySTDinterwordspacing

\bibitem{ma57}
\BIBentryALTinterwordspacing
I.~S. Duff, ``Ma57---a code for the solution of sparse symmetric definite and indefinite systems,'' \emph{ACM Trans. Math. Softw.}, vol.~30, no.~2, pp. 118--144, June 2004. [Online]. Available: \url{https://doi.org/10.1145/992200.992202}
\BIBentrySTDinterwordspacing

\bibitem{Coleman2014}
D.~Coleman, I.~A. Sucan, S.~Chitta, and N.~Correll, ``Reducing the barrier to entry of complex robotic software: a {MoveIt!} case study,'' \emph{Journal of Software Engineering for Robotics}, vol.~5, no.~1, pp. 3--16, 2014.

\bibitem{Kuffner2000}
J.~Kuffner and S.~LaValle, ``{RRT}-connect: An efficient approach to single-query path planning,'' in \emph{Proceedings of the IEEE International Conference on Robotics and Automation}, vol.~2, 2000, pp. 995--1001.

\bibitem{Kunz2012}
T.~Kunz and M.~Stilman, ``Time-optimal trajectory generation for path following with bounded acceleration and velocity,'' in \emph{Proceedings of Robotics: Science and Systems}, 2012, pp. 1--8.

\end{thebibliography}
                                                   







\end{document}